\documentclass[letterpaper]{article} 
\usepackage[preprint]{aaai2027}  
\usepackage[hyphens]{url}  
\usepackage{graphicx} 
\usepackage{natbib}  
\usepackage{caption} 
\usepackage{algorithm}
\usepackage{algorithmic}
\usepackage{multirow}
\usepackage{amsmath,amsfonts}

\usepackage{newfloat}
\usepackage{listings}
\DeclareCaptionStyle{ruled}{labelfont=normalfont,labelsep=colon,strut=off} 
\floatstyle{ruled}
\newfloat{listing}{tb}{lst}{}
\floatname{listing}{Listing}

\usepackage{booktabs}

\title{Latent Action as Intention Enables Efficient Future Imagination\\for World Action Models}
\author{
    Xiang Li\textsuperscript{\rm 1,2,4}\equalcontrib,
    Yupeng Zheng\textsuperscript{\rm 3}\equalcontrib\corresponding\thanks{Project Leader.},
    Songen Gu\textsuperscript{\rm 5}\equalcontrib,
    Huailiang Ma\textsuperscript{\rm 6}\equalcontrib,
    Feng Yu\textsuperscript{\rm 7},
    Yuhang Zheng\textsuperscript{\rm 8},\\
    Xian Nie\textsuperscript{\rm 7},
    Shanshuai Yuan\textsuperscript{\rm 4,5}, Yujie Zang\textsuperscript{\rm 8},
    Weize Li\textsuperscript{\rm 8}, Shuai Tian\textsuperscript{\rm 3},
    Moyang Liu\textsuperscript{\rm 9},\\
    Ya-Qin Zhang\textsuperscript{\rm 2}, Wenchao Ding\textsuperscript{\rm 4}\corresponding
}
\affiliations{
    \textsuperscript{\rm 1}CollegeAI, THU~
    \textsuperscript{\rm 2}AIR, THU~
    \textsuperscript{\rm 3}CASIA~
    \textsuperscript{\rm 4}TARS Robotics~
    \textsuperscript{\rm 5}FDU~
    \textsuperscript{\rm 6}Southeast University~
    \textsuperscript{\rm 7}SJTU~
    \textsuperscript{\rm 8}NUS~
    \textsuperscript{\rm 9}BUAA\\
    Project page: https://getterupper.github.io/LAWA
}

\begin{document}

\maketitle


\begin{abstract}
World action models (WAMs) improve robot control by modeling how observations evolve,
but generating future observations at test time incurs substantial latency.
Fast-WAM removes this process for efficiency;
however, our matched implementations show lower generalization for Fast-WAM
than for future-aware alternatives, especially with scarce robot demonstrations
and in out-of-distribution scenarios.
To bridge this gap, we introduce \textbf{LAWA}, a WAM architecture that
uses compact latent actions as an operational representation of future intentions,
enabling efficient test-time future imagination without generating future observations.
Specifically, a discrete tokenizer
enhanced by action-free pre-training produces
manipulation-centric
codebook targets. LAWA jointly denoises a continuous latent
state anchored to these targets with executable action chunks while omitting
the future-video branch at inference.
On RoboCasa, LAWA achieves state-of-the-art average success rates
of 65.6\% and 80.8\% in the few-shot and full data settings,
improving over the matched Fast-WAM baseline by 9.6 and 4.5 points,
respectively. It also preserves the performance level of the matched Joint-WAM
variant while requiring 42.9\% lower inference latency.
LAWA also demonstrates competitive zero-shot robustness on
LIBERO-Plus and superior performance on real-world tasks.
These results show
that future imagination need not be discarded: retaining it with
compact latent actions yields an effective
trade-off among performance, generalization, and latency.
Code and models will be released.

\end{abstract}

\begin{figure*}[t]
\centering
\includegraphics[width=0.9\linewidth]{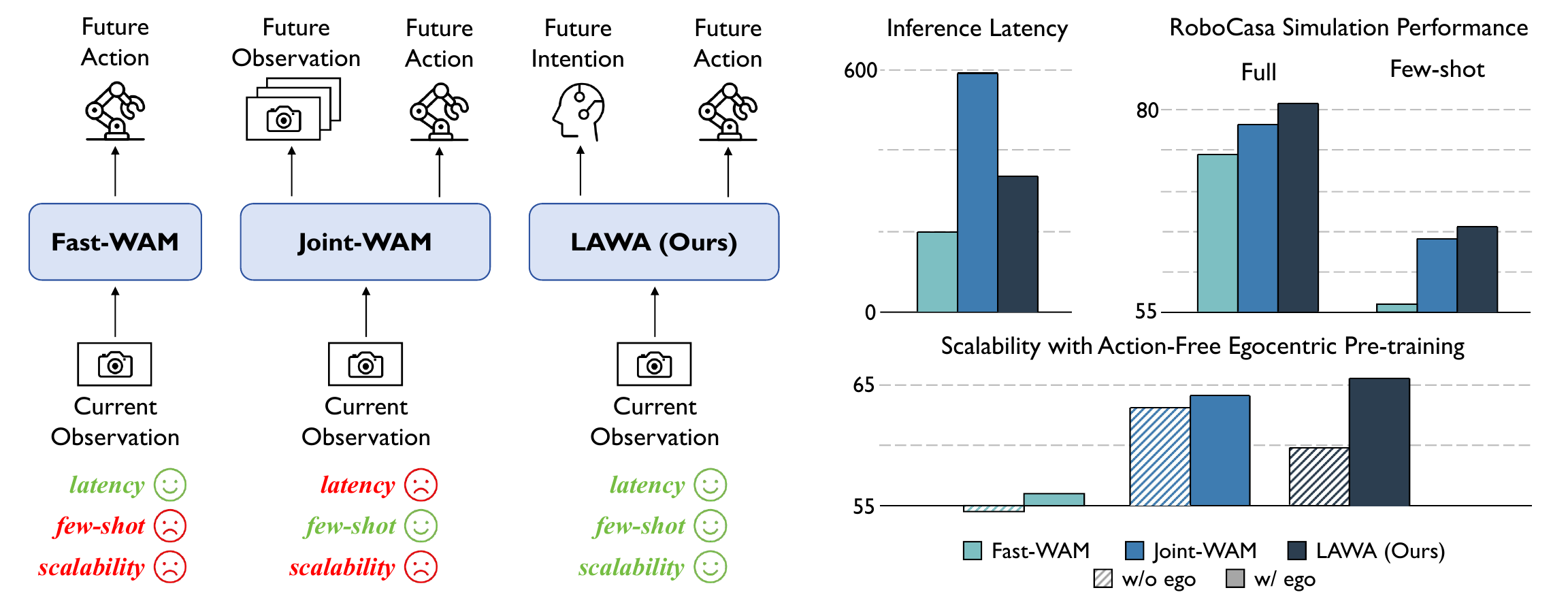} 
\caption{\textbf{Conceptual and quantitative comparison of WAM paradigms.}
Fast-WAM, our primary performance baseline, removes test-time future
imagination but generalizes worse; Joint-WAM predicts future observations and
serves as the efficiency reference. LAWA treats latent actions as future
intentions, substantially
improving performance over Fast-WAM while matching
Joint-WAM at lower latency. Moreover, LAWA obtains larger downstream gains from
action-free data than both alternatives under matched video exposure and
paradigm-specific pre-training objectives.}
\label{fig:teaser}
\end{figure*}

\section{Introduction}
Robotic manipulation requires more than reacting to the current observation:
a policy must also understand how an interaction should unfold. World action
models (WAMs) address this need by coupling action learning with
predictive modeling of future visual
dynamics~\cite{liao2025genie,li2025unified,ma2026dit4dit,yu2026maskwam}.
Predicting future observations
provides the action
generator with an explicit representation of task progress and environmental
dynamics, which is particularly valuable when labeled robot
demonstrations are scarce~\cite{liang2025video,kim2026cosmos}.
This Joint-WAM mechanism~\cite{bi2026motus,ye2026world,li2026causal},
however, creates a practical
bottleneck. Iteratively denoising future observations adds
substantial test-time computation before an action can be executed,
thereby posing an obstacle to real-robot control.

Fast-WAM~\cite{yuan2026fast} challenges the need for test-time future
imagination: it preserves video prediction during training but removes the
future-video branch at inference. Our controlled comparison in
Figure~\ref{fig:teaser}, however, shows that it generalizes markedly worse than
matched Joint-WAM, especially with few-shot supervision.
This result suggests that the benefit of
future modeling may not arise solely from video co-training
and motivates retaining an explicit future representation.
Yet retaining
observation-space imagination incurs the full cost of visual generation, leaving a
trade-off between generalization
performance and inference efficiency in current WAM paradigm designs.

We introduce \textbf{LAWA}, which addresses this imbalance by moving future
imagination from the observation space to a compact latent-action space. Our
central idea is to treat a temporally structured sequence of latent actions as
\emph{future intentions}.
Here, \textit{intention} is an operational term for the predicted sequence of
transition targets available to the action expert.
This sequence encodes
task-relevant transitions that
should occur, without reconstructing all visual details of the future.
LAWA jointly trains video, latent action, and action
denoising experts through multi-model joint attention. A structured attention
mask prevents access to future observations while allowing the action expert to
use the evolving latent intention. At inference, LAWA discards the future-video
branch and jointly denoises only latent intentions and action chunks.
It therefore retains future-aware action generation without performing
expensive visual prediction.

Learning a compact intention space introduces a second challenge: video
reconstruction can favor static appearance over the small interaction dynamics
that determine manipulation success. We address this issue with a discrete
latent action tokenizer trained on action-free robot and egocentric videos.
The tokenizer compresses visual transitions into discrete latent actions,
while a forward decoder provides a next-observation reconstruction objective
during training. An auxiliary mask-prediction objective is designed to bias
the resulting tokens toward hands, manipulators, and their interaction regions.
We automatically generate the mask targets with SAM~2~\cite{ravi2024sam2segmentimage},
thereby enabling scalable annotation-free tokenizer training.
We align
motion-speed distributions across heterogeneous data sources and rebalance
sampling to preserve robot-domain grounding.
On RoboCasa, we empirically find that, without egocentric pre-training,
the latent-action-based policy remains weaker than the matched Joint-WAM with
explicit future-observation prediction,
whereas scalable video pre-training reverses the
gap. These results motivate compact latent actions as an efficient intention
interface and show that action-free egocentric video provides complementary
training diversity.

Experiments across few-shot, full-data, and zero-shot settings support this
account. On RoboCasa, LAWA achieves state-of-the-art 
average success rates of 65.6\% with only
10\% of the training trajectories and 80.8\% with full data, outperforming the
matched Fast-WAM baseline by 9.6 and 4.5
points, respectively,
while preserving the performance level of matched Joint-WAM.
On LIBERO-Plus, LAWA reaches 74.4\% zero-shot success under diverse
perturbations, 14.4 points above the matched Fast-WAM and 4.0
points above Joint-WAM.
On real-world tasks, LAWA demonstrates superior performance on both
fine-grained assembly and long-horizon manipulation,
surpassing full-data Fast-WAM with only 25\% of the demonstrations.
Despite retaining
test-time future imagination, LAWA reduces inference latency by 42.9\% relative
to Joint-WAM while substantially outperforming the faster Fast-WAM.
Controlled
ablations further show larger downstream gains for LAWA from action-free
egocentric pre-training than for both alternatives and continued improvement as
the video corpus scales. This is a matched-video
comparison: the paradigms use the same clips and preprocessing but their native
objectives and trainable modules differ (see \textbf{Appendix}).
Together, these results show that compact latent actions
provide an effective balance among performance, generalization, and
inference efficiency.

In summary, our contributions are as follows.
\begin{itemize}
    \item We introduce LAWA, a WAM that treats compact latent actions as future
    intentions, retaining test-time
    future imagination without generating future observations.
    \item We develop a mask-supervised latent action tokenizer enhanced by
    scalable action-free pre-training.
    \item We demonstrate the effectiveness of our paradigm through extensive experiments
    across simulation benchmarks and real-world tasks,
    substantially improving performance over Fast-WAM while retaining
    Joint-WAM-level performance with lower inference latency.
\end{itemize}

\section{Related Work}

\paragraph{World Action Models.}
World action models (WAMs) augment visuomotor policies with predictive
modeling so that action generation can account for how a scene may
evolve~\cite{zhu2025unified,pai2025mimic,ye2026world,kim2026cosmos}.
Motus jointly models future images and actions~\cite{bi2026motus}, while
DiT4DiT conditions action diffusion on video-denoising
features~\cite{ma2026dit4dit}.
Generating future observations, however, incurs substantial inference cost.
Fast-WAM therefore removes future prediction at test time while retaining
video co-training~\cite{yuan2026fast}, whereas Being-H0.7 aligns
current-conditioned latent queries with a training-only future
posterior~\cite{luo2026being}.
In contrast,
LAWA
retains \emph{test-time future imagination} in a compact space:
predicted latent actions serve as future intentions without future-observation
generation.

\begin{figure*}[t]
\centering
\includegraphics[width=0.9\linewidth]{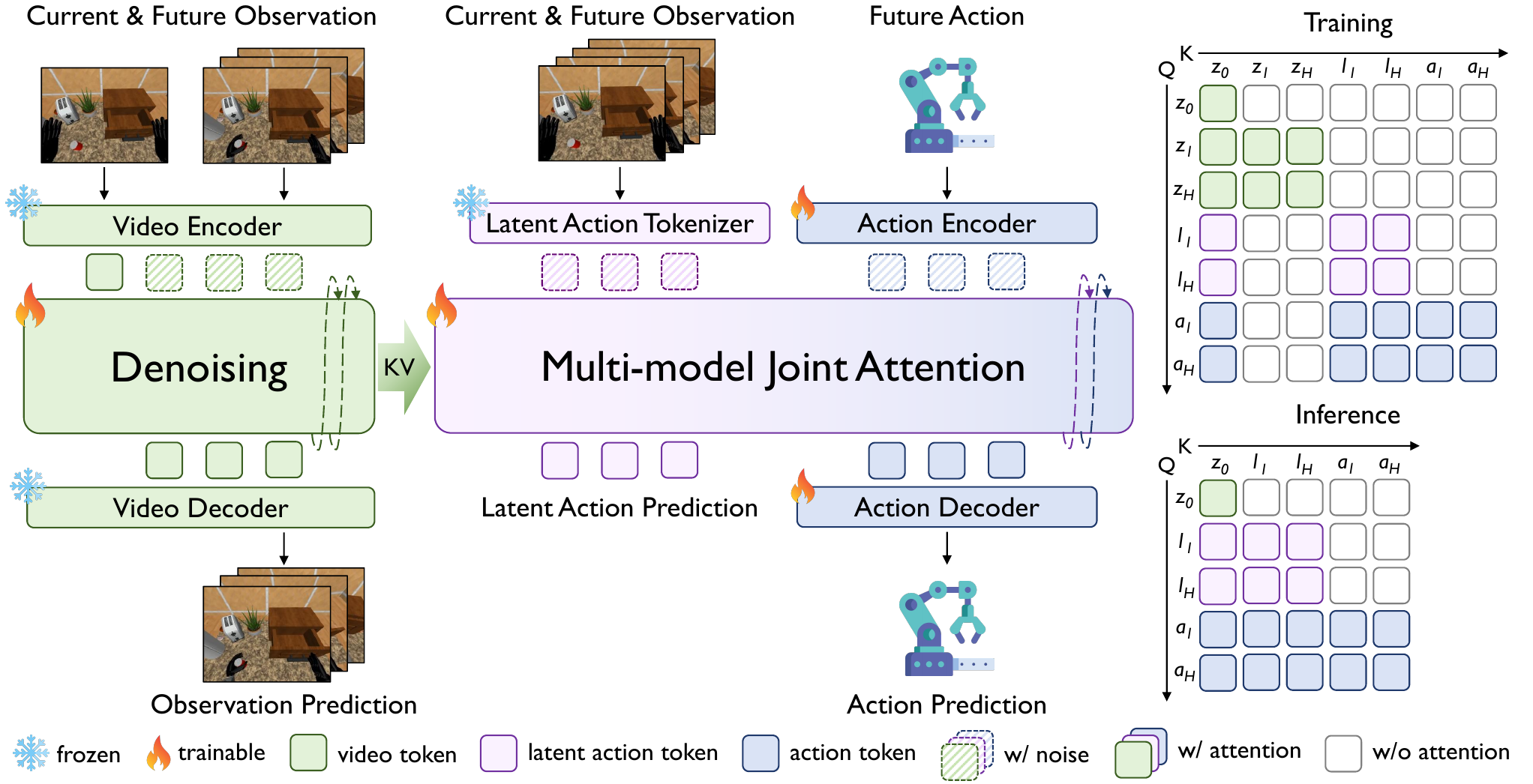} 
\caption{\textbf{Overview of our LAWA.}
A structured mask controls information flow among the three branches.
At inference, LAWA
discards the future-video branch and jointly denoises compact latent intentions
and action chunks, retaining future-aware action generation without explicitly
predicting future observations. The language instruction is omitted for better illustration.}
\label{fig:main}
\end{figure*}

\paragraph{Latent Actions for Robotic Manipulation.}
Latent actions compress visual transitions into action-like variables, enabling
pre-training on videos without action
labels~\cite{schmidt2024learning,liang2025clam,zhang2026disentangled,li2026imagined}.
LAPA quantizes inter-frame transitions into discrete codes for VLA
pre-training and adapts them to robot actions with labeled
demonstrations~\cite{ye2024lapa}.
UniVLA learns task-centric latent actions~\cite{bu2025univla}; ViPRA grounds them via
future-video prediction~\cite{routray2025vipra}; and Motus
leverages optical flow to learn motion-centric latent actions~\cite{bi2026motus}.
DIAL, an end-to-end VLA in our taxonomy, predicts continuous visual foresight
in a VLM space and uses a separate inverse-dynamics
policy~\cite{chen2026dial}. Being-H0.7 transfers future information from a
training-only posterior to deployable latent queries~\cite{luo2026being}.
LAWA instead predicts a ViPRA-style discrete transition sequence online and
jointly denoises its continuous relaxation with executable actions at
test time.
\textbf{Appendix} provides a structured comparison.

\section{Method}

We consider a policy that maps the current observation $o_t$ and language
instruction $c$ to an action chunk $a_{t+1:t+H}$. Fast-WAM~\cite{yuan2026fast} couples a pre-trained video
Diffusion Transformer with an action expert through shared attention. During
training, its branches denoise future-observation latents $z_{t+1:t+H}$ and
predict continuous actions; a structured attention mask prevents
the action tokens from accessing future-video tokens, so both branches learn
from the common current observation tokens $z_t$ without future-information leakage. At inference, Fast-WAM encodes the current observation once and omits
video denoising, gaining speed but losing an explicit representation of how the task should progress.
In contrast to
observation-level future imagination like Joint-WAM,
our LAWA performs latent-action future imagination:
as shown in
Figure~\ref{fig:main}, it predicts a compact
latent action sequence as future intentions and uses it to guide the action
expert,
thereby preserving an efficient inference interface like Fast-WAM.

\begin{figure}[t]
\centering
\includegraphics[width=\linewidth]{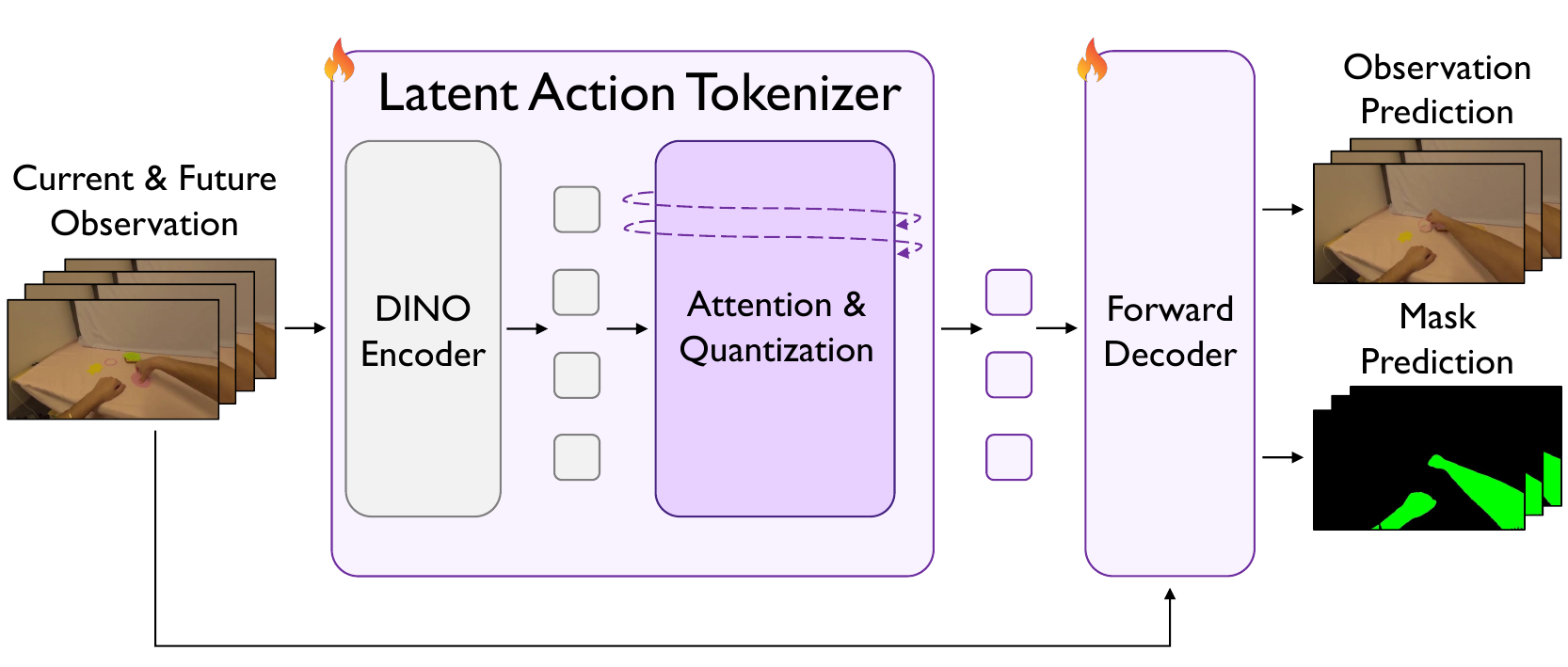} 
\caption{Action-free egocentric pre-training pipeline of the latent action tokenizer.}
\label{fig:lam}
\end{figure}

\subsection{Latent Action Tokenizer}
\paragraph{Architecture.}
Building upon ViPRA~\cite{routray2025vipra},
the tokenizer maps an observation sequence $o_{t:t+H}$ to discrete latent action
tokens $l_{t+1:t+H}$ that capture dynamic transitions without using
action labels. A
DINOv2~\cite{oquab2023dinov2} encoder first extracts patch
tokens for every frame. A non-causal transformer then
contextualizes these tokens through factorized spatial and temporal attention,
producing $F_k\in\mathbb{R}^{N\times d}$ associated with each observation $o_k$.
We encode the transition
by
differencing $F_k$ and $F_{k-1}$, and spatially compress the result into
$L$ tokens $h_k\in\mathbb{R}^{L\times d_q}$. Each compressed token is assigned to its
nearest entry in a learnable codebook $\mathcal{C}=\{e_j\}_{j=1}^{K}$.
For $p=1,\ldots,L$,
\begin{equation}
 j_{k,p}^{\star}=\arg\min_j\|h_{k,p}-e_j\|_2^2,~
 l_{k,p}=e_{j_{k,p}^{\star}}.
\end{equation}
To ground the codes in observable
dynamics, a causal forward decoder concatenates the projected latent action
$l_k$ as additional spatial tokens with the patch tokens of $o_{k-1}$ and
reconstructs $o_k$. Thus, the discrete representation must retain transition
information that cannot be inferred from static appearance alone.

\paragraph{Manipulation-centric Auxiliary Supervision.}
Video reconstruction can be dominated by scene appearance,
causing the tokenizer to underrepresent small but task-critical interactions.
Motus~\cite{bi2026motus} proposes optical flow as a suitable motion representation;
however, we empirically find that mask prediction is more effective.
We therefore augment the video-prediction tokenizer with a SAM-style mask decoder~\cite{kirillov2023segment}.
For each transition, the decoder takes detached visual reconstruction tokens
as image tokens and the projected latent action as a prompt,
and predicts the corresponding hand or robot-manipulator mask.
We automatically generate mask targets using SAM~2~\cite{ravi2024sam2segmentimage},
thus avoiding laborious manual annotation and enabling scalable pre-training.
This objective provides region-level supervision designed to bias the latent
actions toward interaction regions. We evaluate its downstream effect in
Table~\ref{tab:ablation-component}, without assuming that the auxiliary loss
alone determines the semantics of the learned codes.

\paragraph{Action-free Egocentric Pre-training.}
Empirically, we observe that a
tokenizer trained exclusively on robot data can
capture transitions
and provide compact latent action targets for future
imagination,
but the limited data diversity restricts its generalization capabilities.
To address this issue, as shown in Figure~\ref{fig:lam}, we leverage
egocentric pre-training, which co-trains the tokenizer and forward decoder on
action-free videos from both robot data and egocentric
datasets.
We apply per-source frame sampling during pre-training
to align their motion speed distributions. This alignment reduces optimization
interference caused by large differences in execution rates across heterogeneous
sources, as also observed in Qwen-RobotManip~\cite{qwenrobotmanip2026}.
We further apply weighted rebalancing instead of sampling proportionally
to dataset size, so that robot videos constitute approximately 20\% of the expected
training samples. Together, speed alignment and weighted sampling preserve
robot-domain grounding while allowing the model to exploit 
diverse egocentric manipulation videos.
See \textbf{Appendix} for data, optimization, and comparison details.

\subsection{Latent Action as Intention}
During robot-policy training, we freeze the pre-trained tokenizer and use it to
encode $o_{t:t+H}$ into target latent actions $l_{t+1:t+H}$. LAWA jointly trains
video, latent action, and action denoising experts, keeping the current observation
clean while independently corrupting their future targets. At inference, LAWA
omits future-video generation, encodes $o_t$ once, and jointly denoises the latent
actions and executable actions from noise. The predicted latent states are
trained toward codebook-derived transition targets and are used as future
intentions for efficient future imagination.
The tokenizer targets are discrete codebook embeddings, but the latent expert
performs flow matching in their continuous embedding space. We do not apply
nearest-neighbor projection during or after denoising; the action expert
conditions on the evolving continuous latent state. Thus, \textit{discrete} describes
the tokenizer's target set, whereas test-time latent imagination is a continuous
relaxation anchored to that set.

\paragraph{Multi-model Joint Attention.}
We couple the three experts through joint attention while retaining
modality-specific Transformer blocks and timestep conditioning.
As shown in Figure~\ref{fig:main}, through a structured attention mask, current-observation
tokens $z_t$ cannot access the future, whereas noisy future-video tokens attend to all
video tokens. Latent action tokens attend only to the current observation and
latent action sequence; action tokens attend to the current observation, latent
action sequence, and action sequence.
This mask lets
the action expert use the evolving future intention without leaking future-video
information. All experts also condition on language and, when available, current
proprioception. At inference, future-video tokens are omitted and the
current-observation features are cached, so iterative denoising evaluates only the
latent action and action experts. The same visibility mask is maintained across
training and inference, while future-observation generation is avoided at test
time.

\subsection{Training Objective}
Training proceeds in two stages. We first pre-train the latent action tokenizer
on action-free videos using
\begin{equation}
 \mathcal{L}_{\mathrm{tok}}=
 \mathcal{L}_{1}+\lambda_{\mathrm{perc}}\mathcal{L}_{\mathrm{perc}}
 +\lambda_{\mathrm{mask}}\mathcal{L}_{\mathrm{mask}},
\end{equation}
where $\mathcal{L}_{1}$ and the LPIPS-based $\mathcal{L}_{\mathrm{perc}}$
supervise next-frame reconstruction~\cite{zhang2018unreasonable}.
For mask prediction,
we utilize binary cross-entropy, Dice and IoU losses as $\mathcal{L}_{\mathrm{mask}}$.
During training, we adopt noise-substitution quantization~\cite{vali2022nsvq}
to avoid additional codebook loss terms.
Low-usage entries are periodically
refreshed to prevent codebook collapse.

We then freeze the tokenizer and optimize the
experts with flow matching. Let $y^m$ be the clean target for modality
$m\in\{\mathrm{vid},\mathrm{lat},\mathrm{act}\}$, corresponding to future-video
latents, latent actions, and executable action chunks. For each branch, we independently
sample Gaussian noise $\epsilon^m\sim\mathcal{N}(0,I)$ and a shifted flow time
$\tau_m$, construct $y_{\tau_m}^m=(1-\tau_m)y^m+\tau_m\epsilon^m$, and set
$v^m=\epsilon^m-y^m$. Let $\hat v_\theta^m$ denote the predicted velocity
conditioned on $y_{\tau_m}^m$, $\tau_m$, and the context $\mathcal{C}_m$
permitted by multi-model joint attention. We minimize
\begin{equation}
 \mathcal{L}_{m}=\mathbb{E}\!\left[w_m(\tau_m)
 \left\|\hat v_\theta^m-v^m\right\|_2^2\right],
\end{equation}
where $w_m(\tau_m)$ is the scheduler-dependent training weight.
We mask padded targets when computing their respective losses.
The robot-policy training
objective is
\begin{equation}
 \mathcal{L}_{\mathrm{LAWA}}=
 \lambda_{\mathrm{vid}}\mathcal{L}_{\mathrm{vid}}+
 \lambda_{\mathrm{lat}}\mathcal{L}_{\mathrm{lat}}+
 \lambda_{\mathrm{act}}\mathcal{L}_{\mathrm{act}}.
\end{equation}
The video term preserves the world-modeling signal, the latent term matches
codebook-derived transition targets, and the action term learns executable control.

\begin{table}[t]
\centering
\setlength{\tabcolsep}{1mm}
\begin{tabular}{lc|cc}
\toprule
\multirow{2}{*}{\text{Method}} & \multirow{2}{*}{\text{Paradigm}} & \multicolumn{2}{c}{SR (\%)} \\
 & & Few-shot & Full \\
\midrule
GR00T N1.6 (\citeauthor{bjorck2025gr00t}) & VLA & - & 47.6 \\
StarVLA (\citeauthor{community2026starvla}) & VLA & - & 48.8 \\
StarVLA-$\alpha$ (\citeauthor{ye2026starvla}) & VLA & - & 53.8 \\
TwinBrainVLA (\citeauthor{yu2026twinbrainvla}) & VLA & - & 54.6 \\
ABot-M0 (\citeauthor{yang2026abot}) & VLA & - & 58.3 \\
RLDX-1 (\citeauthor{kim2026rldx}) & VLA & - & 58.7 \\
JoyAI-RA (\citeauthor{zhang2026joyai}) & VLA & - & 63.2 \\
DIAL (\citeauthor{chen2026dial}) & VLA & 58.3 & 70.2 \\
\midrule
Being-H0.7 (\citeauthor{luo2026being}) & WAM & - & 49.2 \\
DiT4DiT (\citeauthor{ma2026dit4dit}) & WAM & - & 50.8 \\
LDA-1B (\citeauthor{lyu2026lda}) & WAM & - & 55.4 \\
\midrule
Fast-WAM$\dagger$ (\citeauthor{yuan2026fast}) & WAM & 56.0 & 76.3 \\
Joint-WAM$\dagger$ & WAM & 64.1 & 78.8 \\
\textbf{LAWA (Ours)} & WAM & \textbf{65.6} & \textbf{80.8} \\
\bottomrule
\end{tabular}
\caption{\textbf{Results on RoboCasa benchmark.} Success rates (SR) across
24 tabletop tasks in both few-shot and full data settings.
Best results in \textbf{bold}.
$\dagger$ denotes our implementation.
}
\label{tab:robocasa}
\end{table}

\section{Experiments}
\label{sec:experiments}

\subsection{Experimental Setup}

\paragraph{RoboCasa.}
Following the protocol of DIAL~\cite{chen2026dial}, we evaluate our method on RoboCasa~\cite{nasiriany2024robocasa}, a simulation benchmark containing 24 tabletop rearrangement and articulated-object tasks.
The \emph{full} data setting uses 24,000 trajectories (1,000 per task) for training,
whereas the \emph{few-shot} setting utilizes a 10\% subset. The latter therefore
tests whether a model can benefit from test-time future imagination
when robot supervision is scarce.
We report the average success rates over 50 trials per task.

\paragraph{LIBERO-Plus.}
We further evaluate zero-shot transfer performance on LIBERO-Plus~\cite{fei2026liberoplus}, which
expands LIBERO~\cite{liu2023libero} with challenging environmental perturbations. Following prior work, we report average success rate for
each perturbation across all four task suites and their micro average. All models
are trained solely on the original LIBERO training trajectories
and receive no
fine-tuning on the augmented LIBERO-Plus training data.


\paragraph{Implementation Details.}
Our Fast-WAM and Joint-WAM implementations and LAWA share downstream splits,
optimization, batch sizes, and training steps, but not paradigm-specific
branches, objectives, parameter counts, or inference costs. Fast-WAM is the
primary performance baseline; Joint-WAM measures the cost of explicit future
generation. We measure latency per action-chunk prediction on identical hardware
and defer full details to the \textbf{Appendix}.

\subsection{Performance on Simulation Benchmarks}

\paragraph{Main Results on RoboCasa.}
As shown in Table~\ref{tab:robocasa},
we compare LAWA against several state-of-the-art VLA and WAM methods.
LAWA achieves the best results in both few-shot and full-data settings. It
reaches 65.6\% with 100 demonstrations per task and 80.8\% with full data,
exceeding matched Fast-WAM by 9.6 and 4.5 points and attaining comparable
success to Joint-WAM (64.1\% and 78.8\%). The two comparisons establish the
complete system's performance gain and its retention of
observation-imagination performance, respectively.
LAWA also exceeds DIAL by 7.3 points in few-shot,
surpassing all reported WAMs and most full-data VLA baselines. Together with
the studies below, these results support using a compact latent sequence as a future-intention
interface without computationally expensive future-observation generation.

\paragraph{Zero-shot Transfer Results on LIBERO-Plus.}
As shown in Table~\ref{tab:libero-plus}, LAWA achieves a competitive
overall success rate
of 74.4\% on LIBERO-Plus, surpassing OpenVLA-OFT by 4.8 points
and matched Joint-WAM's 70.4\%.
Under the aligned training recipe, our matched Fast-WAM implementation reaches
60.0\%, whereas LAWA reaches 74.4\%, outperforming it by 14.4 points and achieving higher success
across all six non-linguistic
observation- and proprioceptive-level perturbations. The gains are especially
pronounced under camera-viewpoint and sensor-noise shifts, reaching 44.3 and
27.5 points, respectively. This controlled system-level gap is consistent with
the value of retaining a test-time future representation under distribution
shift. LAWA realizes that representation with compact latent actions rather
than explicit future observations.

\begin{table*}[t]
\centering
\setlength{\tabcolsep}{1mm}
\begin{tabular}{lc|ccccccc|c}
\toprule
\multirow{2}{*}{\text{Method}} & \multirow{2}{*}{\text{Paradigm}} & \multicolumn{7}{c|}{Perturbation Type} & \multirow{2}{*}{\text{Total}} \\
 & & Camera & Robot & Language & Light & Background & Noise & Layout & \\
\midrule
OpenVLA (\citeauthor{kim2024openvla}) & VLA & 0.8 & 3.5 & 23.0 & 8.1 & 34.8 & 15.2 & 28.5 & 15.6 \\
OpenVLA-OFT (\citeauthor{kim2025fine}) & VLA & 56.4 & 31.9 & 79.5 & 88.7 & \textbf{93.3} & 75.8 & 74.2 & 69.6 \\
UniVLA (\citeauthor{bu2025univla}) & VLA & 1.8 & 46.2 & 69.6&  69.0 & 81.0&  21.2&  31.9 & 42.9 \\
WorldVLA (\citeauthor{cen2025worldvla}) & VLA & 0.1 & 27.9&  41.6 & 43.7&  17.1 & 10.9 & 38.0 & 25.0 \\
$\pi_0$ (\citeauthor{black2024pi0}) & VLA & 13.8 & 6.0 & 58.8 & 85.0 & 81.4 & 79.0 & 68.9 & 53.6 \\
$\pi_{0}$-FAST (\citeauthor{pertsch2025fast}) & VLA & 65.1 & 21.6 & 61.0 & 73.2 & 73.2 & 74.4 & 68.8 & 61.6 \\
\midrule
Fast-WAM$\dagger$ (\citeauthor{yuan2026fast}) & WAM & 24.9 & 50.7 & 76.9 & 89.2 & 62.3 & 58.0 & 67.7 & 60.0 \\
Joint-WAM$\dagger$ & WAM & 47.2 & 65.3 & \textbf{91.8} & 94.8 & 57.6 & 61.0 & \textbf{78.9} & 70.4 \\
\textbf{LAWA (Ours)} & WAM & \textbf{69.2} & \textbf{66.7} & 62.8 & \textbf{96.2} & 64.5 & \textbf{85.5} & 78.6 & \textbf{74.4} \\
\bottomrule
\end{tabular}
\caption{\textbf{Zero-shot transfer results on LIBERO-Plus benchmark.} Success rates under seven perturbations and their micro average. Best
results in \textbf{bold}.
$\dagger$ denotes our implementation.}
\label{tab:libero-plus}
\end{table*}

\subsection{Comparison between Paradigms}

\begin{figure}[t]
\centering
\includegraphics[width=\linewidth]{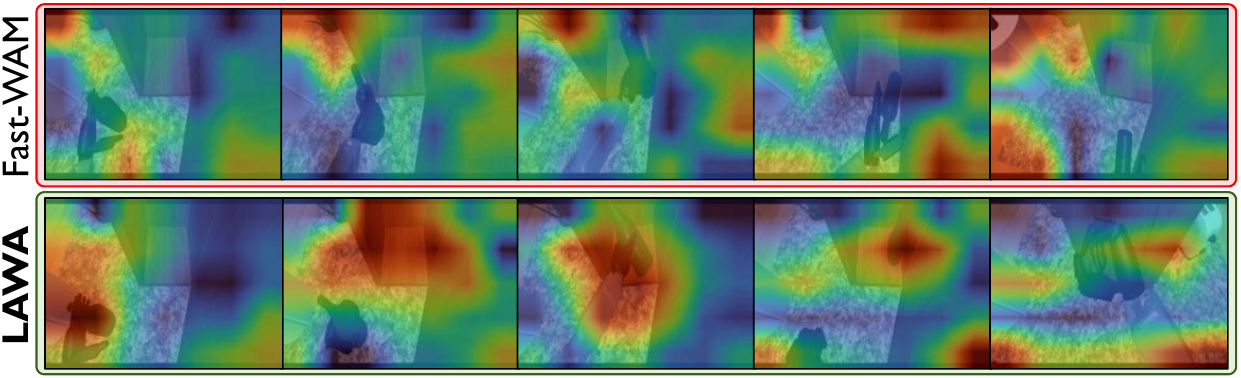}
\caption{\textbf{Action-to-vision attention map comparison.}
In this illustrative rollout, LAWA concentrates more on
the manipulated object and interaction region,
rather than diffuse background appearance.}
\label{fig:attention}
\end{figure}

\begin{table}[t]
\centering
\setlength{\tabcolsep}{1mm}
\begin{tabular}{lccc}
\toprule
Input & Unperturbed & Gaussian &
Temporal Shuffle \\
\midrule
SR (\%) & \textbf{80.8} & 52.2 & 56.4 \\
\bottomrule
\end{tabular}
\caption{\textbf{Inference-time latent-action perturbation results.} Gaussian denotes
isotropic noise with $\sigma=1.0$.}
\label{tab:latent-action-perturbation}
\end{table}

\paragraph{The Effectiveness of Latent Action.}
We first test whether the executor functionally uses the predicted latent
sequence by perturbing only its model-visible latent-action state at inference.
As shown
in Table~\ref{tab:latent-action-perturbation}, Gaussian noise
reduces full-data RoboCasa success from 80.8\% to 52.2\%,
while temporal shuffling yields 56.4\%.
These interventions demonstrate that action generation depends on both
the content and temporal organization of the latent sequence, rather than
ignoring this pathway. Figure~\ref{fig:attention} provides complementary
qualitative evidence on RoboCasa \emph{WineToCabinetClose}.
Fast-WAM distributes attention broadly over the countertop and background and
ultimately fails, whereas LAWA tracks the manipulated object and task-relevant
region across successive interaction stages and completes the task. Together,
the intervention and visualization show that temporally structured latent
actions provide a functionally used, manipulation-focused conditioning signal
for action generation.

\paragraph{The Effectiveness of Egocentric Pre-training.}
Table~\ref{tab:ablation-ego} shows that action-free egocentric
pre-training benefits all three paradigms, but to markedly different degrees.
Fast-WAM improves by 1.5 and 1.7 points and Joint-WAM by only 1.0 and 0.5 points in the
few-shot and full data settings, whereas LAWA gains 5.9 and 4.5 points, respectively.
Notably, without
egocentric pre-training, LAWA does not yet match Joint-WAM: it trails by 3.4 and
2.0 points.
Thus, compact latent-action prediction alone is
not uniformly stronger than explicit future-observation prediction. After pre-training,
however, LAWA reaches 65.6\% and 80.8\% on RoboCasa, matching the Joint-WAM
performance reference while retaining its inference-efficiency advantage.
Under this matched-video protocol, each paradigm uses its native objective and
trainable modules.
The results therefore establish larger downstream gains for LAWA under
the stated schedules.
A plausible
explanation is that the compact
bottleneck reduces appearance shortcuts while mask supervision emphasizes
interaction regions (see Table~\ref{tab:ablation-component}).

\begin{table}[t]
\centering
\begin{tabular}{l|cc|cc}
\toprule
\multirow{2}{*}{\text{Method}} & \multicolumn{2}{c|}{Few-shot SR (\%)} & \multicolumn{2}{c}{Full SR (\%)} \\
 & w/o ego & w/ ego & w/o ego & w/ ego \\
\midrule
Fast-WAM & 54.5 & 56.0 & 74.6 & 76.3 \\
Joint-WAM & \textbf{63.1} & 64.1 & \textbf{78.3} & 78.8 \\
\textbf{LAWA (Ours)} & 59.7 & \textbf{65.6} & 76.3 & \textbf{80.8} \\
\bottomrule
\end{tabular}
\caption{Egocentric pre-training effectiveness comparison.}
\label{tab:ablation-ego}
\end{table}

\begin{table}[t]
\centering
\begin{tabular}{l|ccc}
\toprule
Method & Fast-WAM & Joint-WAM & LAWA \\
\midrule
Latency (ms) & 196.5 & 593.1 & 338.5 \\
\bottomrule
\end{tabular}
\caption{\textbf{End-to-end inference latency comparison.} Results measured on one
NVIDIA A800 GPU.}
\label{tab:ablation-latency}
\end{table}

\begin{figure}[t]
\centering
\includegraphics[width=\linewidth]{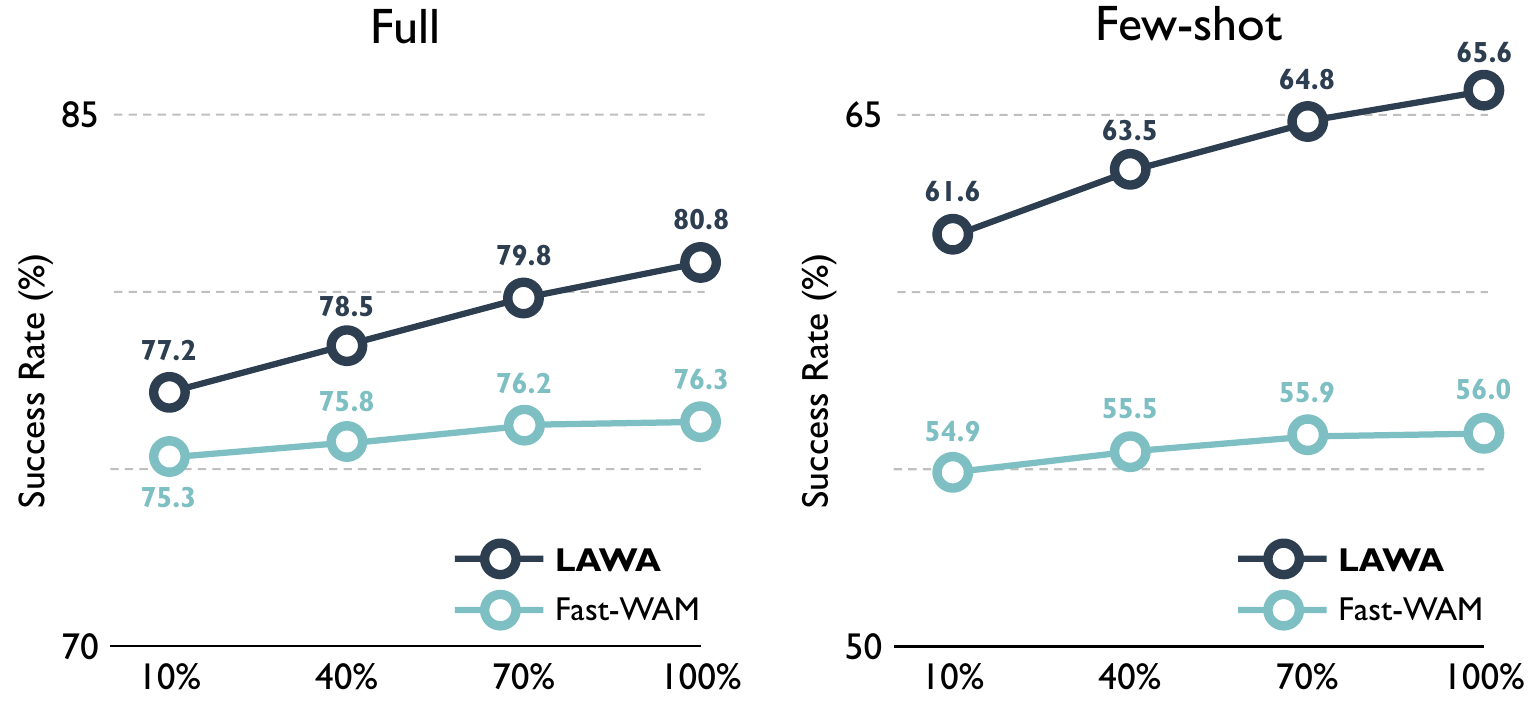}
\caption{\textbf{Egocentric pre-training scalability comparison.} Under
matched video fractions, LAWA obtains larger
downstream gains than Fast-WAM in both data settings.}
\label{fig:scaling}
\end{figure}

\paragraph{The Scalability of Egocentric Pre-training.}
Figure~\ref{fig:scaling} reveals a clear difference in scalability. Increasing
the fraction of action-free egocentric videos from 10\% to 100\% raises LAWA
from 77.2\% to 80.8\% in full-data and from 61.6\% to 65.6\% in
few-shot setting, corresponding to gains of 3.6 and 4.0 points.
Under the corresponding pre-training schedule, Fast-WAM improves by only 1.0
and 1.1 points. The widening downstream gap shows that LAWA continues to benefit
as the video corpus grows, with the largest gain when labeled robot
demonstrations are scarce.

\paragraph{Performance--latency Trade-off.}
As shown in Table~\ref{tab:ablation-latency},
LAWA requires 338.5\,ms per action chunk, 42.9\% below Joint-WAM's 593.1\,ms
at comparable success.
Fast-WAM is faster at 196.5\,ms, but this
advantage
comes with substantially lower success rates.
LAWA therefore improves performance and generalization over
Fast-WAM, while preserving Joint-WAM-level performance at
lower
latency. This complementary comparison defines the favorable
performance--latency trade-off of latent-action future imagination.

\subsection{Ablation Study}
We performed ablation studies of the proposed components
in Table~\ref{tab:ablation-component} by
progressively adding latent actions (LA), egocentric pre-training (EP), and an
auxiliary loss (AL). Introducing latent actions alone improves the Fast-WAM baseline
by 5.2 and 1.7 percentage points in the few-shot and full data settings, respectively.
Action-free egocentric pre-training contributes a further 5.1 and 3.0 points.
These gains show that the latent branch and egocentric pre-training provide
complementary system-level improvements.
Among
the auxiliary targets, we empirically find mask prediction to be more effective
than optical flow prediction. Relative to using no auxiliary loss, flow prediction
degrades performance by 1.3 and 0.7 points in the two settings, whereas mask
prediction improves performance by 0.8 and 1.5 points. Overall, the component
additions are empirically complementary, and the mask-supervised tokenizer
yields a further modest improvement. We interpret the mask objective as a
manipulation-oriented inductive bias, rather than direct proof of the learned
representation's semantics.

\begin{table}[t]
\centering
\begin{tabular}{ccc|cc}
\toprule
\multirow{2}{*}{\text{LA}} & \multirow{2}{*}{\text{EP}} & \multirow{2}{*}{\text{AL}} & \multicolumn{2}{c}{SR (\%)} \\
 & & & Few-shot & Full \\
\midrule
 & & & 54.5 & 74.6 \\
\midrule
 \checkmark & & & 59.7 & 76.3 \\
 \checkmark & \checkmark & & 64.8 & 79.3 \\
 \checkmark & \checkmark & Flow & 63.5 & 78.6 \\
 \checkmark & \checkmark & Mask & \textbf{65.6} & \textbf{80.8} \\
\bottomrule
\end{tabular}
\caption{Component ablation results, including Latent Action (LA), Egocentric Pre-training (EP) and Auxiliary Loss (AL).}
\label{tab:ablation-component}
\end{table}

\subsection{Performance on Real-World Tasks}
\label{sec:real-world}

\begin{figure}[t]
\centering
\includegraphics[width=\linewidth]{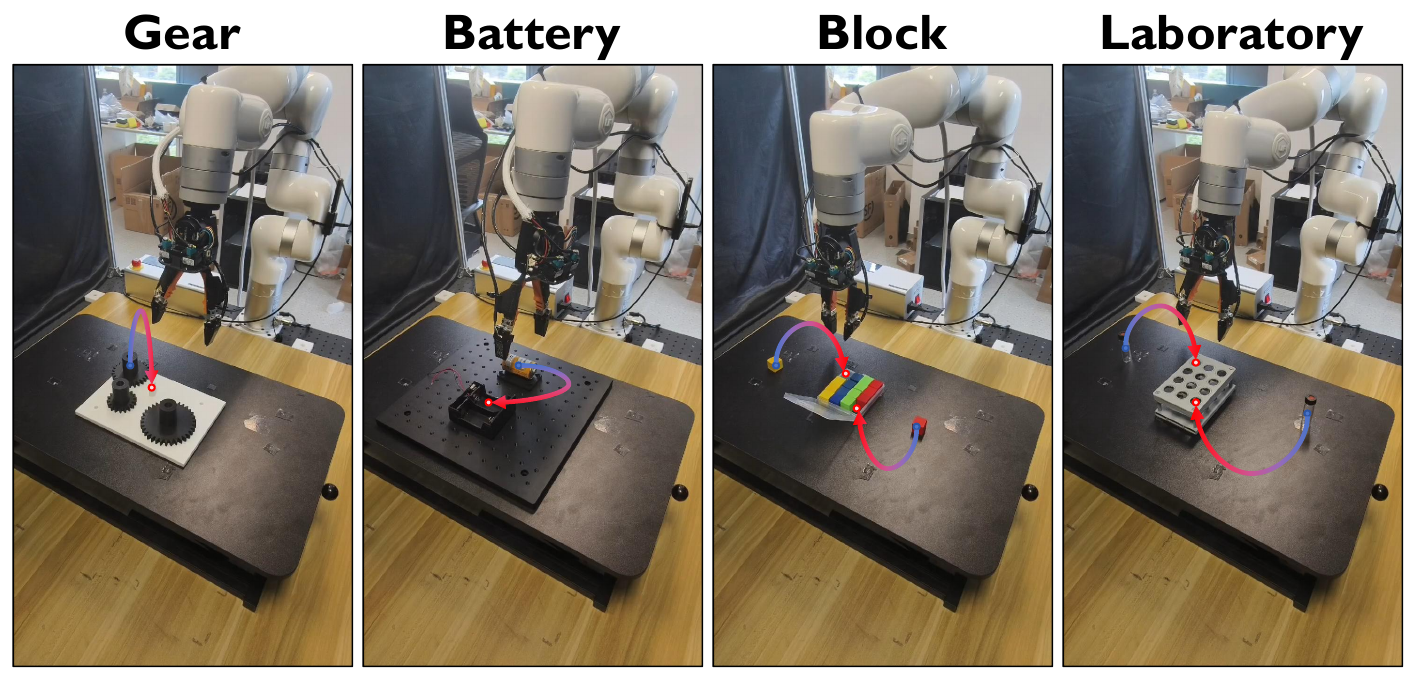}
\caption{
\textbf{Real-world tasks.}
\textit{Gear} and \textit{Battery} involve assembly with
fine-grained operations, while
\textit{Block} and \textit{Laboratory} require long-horizon
manipulation. Colored arrows indicate the required object
transfers from their initial locations (blue) to their target
locations (red).}
\label{fig:setup}
\end{figure}

\begin{table}[t]
\centering
\setlength{\tabcolsep}{1.2mm}
\begin{tabular}{lc|cccc|c}
\toprule
Method & Data & Gear & Batt. & Block & Lab. & Avg. \\
\midrule
Fast-WAM & 25\% & 30.0 & 5.0 & 0.0 & 0.0 & 8.8 \\
\textbf{LAWA (Ours)} & 25\% & \textbf{55.0} & \textbf{30.0} & \textbf{45.0} & \textbf{30.0} & \textbf{40.0} \\
\midrule
Fast-WAM & 50\% & 50.0 & 20.0 & 10.0 & 0.0 & 20.0 \\
\textbf{LAWA (Ours)} & 50\% & \textbf{70.0} & \textbf{50.0} & \textbf{60.0} & \textbf{45.0} & \textbf{56.3} \\
\midrule
Fast-WAM & 100\% & 65.0 & 40.0 & 25.0 & 5.0 & 33.8 \\

\textbf{LAWA (Ours)} & 100\% & \textbf{85.0} & \textbf{65.0} & \textbf{70.0} & \textbf{50.0} & \textbf{67.5}\\
\bottomrule
\end{tabular}
\caption{\textbf{Results on real-world tasks.}
Success rates across each task under different
proportions of robot demonstration trajectories.
Best results of each data setting in \textbf{bold}.
}
\label{tab:real-world}
\end{table}

\paragraph{Setup and tasks.}
We conduct real-world evaluation using a
UFACTORY xArm7 equipped with a gripper.
Besides, we use a RealSense D435 camera providing the base view
and two fisheye cameras providing wrist views
for visual inputs.
We design four real-world tasks to cover complementary manipulation
requirements, as illustrated in Figure~\ref{fig:setup}. \textit{Gear} and
\textit{Battery} are assembly tasks that require relatively fine-grained
operations and are therefore more challenging. For example, the robot must
adjust the gear orientation so that its teeth mesh with those of the adjacent
gear, or press the battery until it is fully seated in the slot. In contrast,
\textit{Block} and \textit{Laboratory} involve multi-stage
manipulation over long
horizons. For each task, we collect 200 demonstration trajectories for training
and evaluate each policy over 20 trials.

\paragraph{Results.}
Table~\ref{tab:real-world} shows that LAWA consistently outperforms Fast-WAM
across all four tasks and all data settings. With 25\%, 50\%, and 100\% of the
demonstrations, LAWA improves over the Fast-WAM baseline by 31.2, 36.3, and 33.8 points,
respectively.
The largest gains in the full-data setting occur on the long-horizon tasks,
where LAWA outperforms Fast-WAM by 45 points on both \textit{Block} and \textit{Laboratory},
consistent with the hypothesis that LAWA's test-time latent pathway benefits
multi-stage execution.
Moreover, LAWA also performs strongly on the fine-grained assembly tasks,
surpassing Fast-WAM by 20 and 25 points on \textit{Gear} and \textit{Battery},
suggesting that the mask-supervised tokenizer may provide useful guidance for
action generation.

LAWA remains effective when robot demonstrations are scarce. Using only 25\%
of the data, corresponding to
50 trajectories per task, LAWA attains an average
success rate of 40.0\%, surpassing the 33.8\% achieved by Fast-WAM with the full
training set. The advantage is particularly pronounced on long-horizon
manipulation: Fast-WAM records no successful trial on either \textit{Block} or
\textit{Laboratory} in this few-shot setting, whereas LAWA reaches 45.0\% and
30.0\%, respectively. Together with Table~\ref{tab:ablation-ego}, these results
show that the complete LAWA system benefits from diverse action-free egocentric
videos when labeled robot demonstrations are limited.
See \textbf{Appendix} for qualitative comparisons.

\section{Conclusion}

We presented LAWA, a WAM that treats compact latent actions operationally as
future intentions and jointly denoises them with executable actions, preserving
future-aware control without observation prediction. Automatic mask supervision
and action-free robot and egocentric videos enhance its discrete tokenizer. Across
RoboCasa, LIBERO-Plus, and real-world tasks, LAWA improves over Fast-WAM while
retaining Joint-WAM-level performance with 42.9\% lower latency.
Unlike Joint-WAMs that explicitly generate future observations, this compact
pathway preserves future-aware action generation without costly visual
prediction. These results show that future imagination need not be discarded
for efficiency, but can instead be represented through compact latent actions.
We hope LAWA establishes a practical WAM paradigm for efficient future
imagination and scalable learning from action-free videos.

\bibliography{aaai2027}


\end{document}